%% file: ms.tex
\documentclass[11pt,a4paper]{article}
\RequirePackage{import}

\import{}{preamble.tex}

\begin{document}
\maketitle

\import{}{body.tex}

\import{}{bibliography.tex}
\end{document}

%% file: preamble.tex
\import{preamble/}{header.tex}
\import{preamble/}{glossaries.tex}
\import{preamble/}{math.tex}

%% file: preamble/header.tex
%
%

\usepackage[hyperref]{emnlp-ijcnlp-2019}
\usepackage{times}
\usepackage{latexsym}
\usepackage[acronym,smallcaps,nowarn,section,nonumberlist]{glossaries}
\glsdisablehyper
\usepackage{url}
\usepackage{booktabs}
\usepackage{graphicx}
\usepackage{cleveref}

\aclfinalcopy 


\title{Neural Language Priors}

\author{
  \parbox{0.9\textwidth}{
    \centering
    Joseph Enguehard \hfill
    Dan Busbridge \hfill
    Vitalii Zhelezniak \hfill 
    Nils Hammerla \\ \vspace{0.1cm}
    {\tt{firstname.lastname@babylonhealth.com}} \\ \vspace{0.1cm}
    Babylon Health, 60 Sloane Ave, Chelsea, London SW3 3DD
  }
}

\date{}

%% file: preamble/glossaries.tex
\makeglossaries

\loadglsentries{preamble/acronyms/a}
\loadglsentries{preamble/acronyms/b}
\loadglsentries{preamble/acronyms/c}
\loadglsentries{preamble/acronyms/d}
\loadglsentries{preamble/acronyms/e}
\loadglsentries{preamble/acronyms/g}
\loadglsentries{preamble/acronyms/h}
\loadglsentries{preamble/acronyms/k}
\loadglsentries{preamble/acronyms/i}
\loadglsentries{preamble/acronyms/l}
\loadglsentries{preamble/acronyms/m}
\loadglsentries{preamble/acronyms/n}
\loadglsentries{preamble/acronyms/o}
\loadglsentries{preamble/acronyms/p}
\loadglsentries{preamble/acronyms/r}
\loadglsentries{preamble/acronyms/s}
\loadglsentries{preamble/acronyms/t}
\loadglsentries{preamble/acronyms/v}
\loadglsentries{preamble/acronyms/w}

%% file: preamble/acronyms/a.tex
\newacronym{argat}{ARGAT}{Across-Relation Graph Attention}
\newacronym{auc}{AUC}{Area Under the Curve}

%% file: preamble/acronyms/b.tex
\newacronym{bow}{BOW}{Bag of Words}
\newacronym{boe}{BOE}{Bag of Embeddings}
\newacronym{borep}{BOREP}{Bag of Random Embedding Projections}
\newacronym{bn}{BN}{Bayesian Network}

%% file: preamble/acronyms/c.tex
\newacronym{cdf}{CDF}{Cumulative Distribution Function}
\newacronym{cnn}{CNN}{Convolutional Neural Network}
\newacronym{cvae}{CVAE}{Conditional Variational Autoencoder}

%% file: preamble/acronyms/d.tex
\newacronym{dag}{DAG}{Directed Acyclic Graph}
\newacronym{dgm}{DGM}{Directed Graphical Model}
\newacronym{dnn}{DNN}{Deep Neural Network}

%% file: preamble/acronyms/e.tex
\newacronym{ecfp4}{ECFP4}{Extended Connectivity Fingerprint}
\newacronym{esn}{ESN}{Echo State Network}

%% file: preamble/acronyms/g.tex
\newacronym{gat}{GAT}{Graph Attention Network}
\newacronym{gcn}{GCN}{Graph Convolutional Network}
\newacronym{gdl}{GDL}{Geometric Deep Learning}
\newacronym{ggnn}{GGNN}{Gated Graph Neural Network}
\newacronym{glove}{GloVe}{Global Vectors for Word Representation}
\newacronym{gnn}{GNN}{Graph Neural Network}
\newacronym{gru}{GRU}{Gated Recurrent Unit}

%% file: preamble/acronyms/h.tex
\newacronym{hmm}{HMM}{Hidden Markov Model}

%% file: preamble/acronyms/k.tex
\newacronym{kld}{KLD}{Kullback-Leibler Divergence}
\newacronym{knn}{KNN}{K-Nearest Neighbour}

%% file: preamble/acronyms/i.tex
\newacronym{idf}{IDF}{Inverse Document Frequency}
\newacronym{ig}{IG}{Information Gain}
\newacronym{isan}{ISAN}{Input Switched Affine Network}

%% file: preamble/acronyms/l.tex
\newacronym{lhs}{L.H.S.}{left hand side}
\newacronym{lstm}{LSTM}{Long Short Term Memory}

%% file: preamble/acronyms/m.tex
\newacronym{map}{MAP}{Maximum a Posteriori}
\newacronym{mdl}{MDL}{Minimum Description Length}
\newacronym{ml}{ML}{Machine Learning}
\newacronym{mle}{MLE}{Maximum Likelihood Estimation}
\newacronym{mnist}{MNIST}{Modified National Institute of Standards and Technology}

%% file: preamble/acronyms/n.tex
\newacronym{nlp}{NLP}{Natural Language Processing}
\newacronym{nn}{NN}{Neural Network}

%% file: preamble/acronyms/o.tex
\newacronym{oh}{OH}{One-Hot}

%% file: preamble/acronyms/p.tex
\newacronym{pdf}{PDF}{Probability Density Function}
\newacronym{pgm}{PGM}{Probabilistic Graphical Model}

%% file: preamble/acronyms/r.tex
\newacronym{relu}{ReLU}{Rectified Linear Unit}
\newacronym{rdf}{RDF}{Resource Description Framework}
\newacronym{rgb}{RGB}{Red Green Blue}
\newacronym{rgc}{RGC}{Relational Graph Convolution}
\newacronym{rgat}{RGAT}{Relational Graph Attention Network}
\newacronym{rgcn}{RGCN}{Relational Graph Convolutional Network}
\newacronym{rhs}{R.H.S.}{Right Hand Side}
\newacronym{rnn}{RNN}{Recurrent Neural Network}
\newacronym{roc}{ROC}{Receiver Operating Characteristic}

%% file: preamble/acronyms/s.tex
\newacronym{sgd}{SGD}{Stochastic Gradient Descent}
\newacronym{sg}{SG}{SkipGram}
\newacronym{st}{ST}{SkipThought}
\newacronym{sts}{STS}{Semantic Textual Similarity}

%% file: preamble/acronyms/t.tex
\newacronym{termfreq}{TF}{Term Frequency}
\newacronym{tfidf}{TF-IDF}{Term Frequency-Inverse Document Frequency}

%% file: preamble/acronyms/v.tex
\newacronym{vae}{VAE}{Variational Autoencoder}

%% file: preamble/acronyms/w.tex
\newacronym{wirgat}{WIRGAT}{Within-Relation Graph Attention}
\newacronym{wl}{WL}{Weisfeiler-Lehman}

%% file: preamble/math.tex
\usepackage{amsmath}
\usepackage{amssymb}
\usepackage{bm}

\input{preamble/math/mathbold}

\input{preamble/math/boldmath}

\input{preamble/math/mathcal}

\input{preamble/math/blackboard}

\input{preamble/math/text}

\input{preamble/math/mathsf}

%% file: preamble/math/mathbold.tex
\newcommand{\mathbold}[1]{\ensuremath{\boldsymbol{\mathbf{#1}}}}

\newcommand{\mbe}{\mathbold{e}}

\newcommand{\mbh}{\mathbold{h}}

\newcommand{\mbk}{\mathbold{k}}

\newcommand{\mbq}{\mathbold{q}}

\newcommand{\mbv}{\mathbold{v}}

\newcommand{\mbE}{\mathbold{E}}

\newcommand{\mbH}{\mathbold{H}}

\newcommand{\mbtheta}{\mathbold{\theta}}

%% file: preamble/math/boldmath.tex
\newcommand{\bme}{\bm{e}}

%% file: preamble/math/mathcal.tex
\usepackage{dutchcal}

%% file: preamble/math/blackboard.tex
\newcommand{\mbbR}{\mathbb{R}}

%% file: preamble/math/text.tex
\newcommand{\txtw}{\text{w}}

\newcommand{\txtS}{\text{S}}
\newcommand{\txtT}{\text{T}}

%% file: preamble/math/mathsf.tex
\newcommand{\bmsfW}{\boldsymbol{\mathsf{W}}}

%% file: body.tex
\import{body/}{abstract.tex}
\import{body/}{introduction.tex}
\import{body/}{method.tex}
\import{body/}{results.tex}
\import{body/}{conclusion.tex}

\import{body/}{acknowledgements.tex}

%% file: body/abstract.tex
\begin{abstract}
  The choice of sentence encoder architecture reflects assumptions about how a
  sentence’s meaning is composed from its constituent words.
  We examine the contribution of these architectures by holding them randomly
  initialised and fixed, effectively treating them as as hand-crafted
  language priors, and evaluating the resulting sentence encoders on downstream
  language tasks. We find that even when encoders are presented with additional
  information that can be used to solve tasks, the corresponding priors do not
  leverage this information, except in an isolated case.
  We also find that apparently uninformative priors are just as good as seemingly
informative priors on almost all tasks, indicating that learning is a necessary
component to leverage information provided by architecture choice.
\end{abstract}

%% file: body/introduction.tex
\section{Introduction}
\label{sec:introduction}
Sentence representations are fixed-length vectors that encode
sentence properties and allow models to learn across many \gls{nlp}
tasks.
These representations enable learning procedures to focus on the training signal
from specific ``downstream'' \gls{nlp} tasks \cite{Conneau}, circumventing the often limited
amount of labelled data.
Naturally, sentence representations that can
effectively encode semantic and syntactic properties into a representations are
highly sought after, and are a cornerstone of modern \gls{nlp} systems.

In practice, sentence representations are formed by applying an encoding function
(or encoder) provided by a \gls{nn} architecture, to the word vectors of the
corresponding sentence. Encoders have been successfully trained to predict
the context of sentence \cite{Kiros2015,Ba2016}, or to leverage supervised
multi-task objectives \cite{Conneau2017b,Dehghani2018a}.

The choice of encoder architecture asserts an \emph{inductive bias}
\cite{Authors2018}, and reflects assumptions about the data-generating process.
Different encoders naturally prioritise one solution over another
\cite{Mitchell1980}, independent of the observed data, trading sample complexity
for flexibility \cite{Geman2008}. Given that \glspl{nn}, which are able to
generalise well, can also overfit when presented with random labels
\cite{Zhang2016a}, we expect that architecture plays a dominant role
in generalisation capability \cite{Lempitsky2018}.

The inductive biases of encoder architectures reflect assumptions about how a
sentence's meaning is composed from its constituent words.
A plethora of architectures have been investigated, each designed with a
specific set of inductive biases in mind. \gls{boe} architectures disregard word
order \cite{Harris1954,Salton1975,Manning2008}, \gls{rnn} architectures can
leverage word positional information \cite{Kiros2015,Ba2016}, \gls{cnn}
architectures compose information at the $n$-gram level
\cite{Collobert2011,Vieira2017,Gan2017}, self-attention models leverage explicit
positional information with long range context \cite{Vaswani2017, Ahmed2017a, Shaw2018,
  Dehghani2018a, Radford2019, Devlin2018, Cer2018a}, and graph-based
models can exploit linguistic structures extracted by traditional \gls{nlp}
methods \cite{Tai2015, Li, Zhang, Teng2016, Kim2018, Ahmed2019,
  Bastings2017,Marcheggiani2017a,Marcheggiani2018,Marcheggiani2018a}. This list
is far from exhaustive.

Given the critical role of encoder architectures in \gls{nlp}, we set out
to examine their contribution to downstream task
performance independent of biases induced by learning processes.
We find that even architectures expected to have extremely strong language
priors yield almost no gains when compared to architectures that are
equipped with apparently uninformative priors, consistent with the results found
in \citet{Wieting2019}.
This suggests that for \gls{nlp} tasks, relying on the prior is insufficient, and the learning process
is necessary, in contrast to what was found in the vision field
\cite{Lempitsky2018}.
In short, although there are known strong inductive biases for language, there
is no best language prior, and in practice there is surprisingly little
correspondence between the two.

To show this, given a set of pre-trained word embeddings, we evaluate the
classification accuracy of a variety architectures on a set of \gls{nlp} tasks,
only updating the parameters specific to the task, holding the parameters of the
architecture fixed at their random initialisation.

%% file: body/method.tex
\section{Method}
\label{sec:method}
\input{body/method/sentence-encoders}

\input{body/method/existing}

\input{body/method/convolutional}
\input{body/results/max-pool-diag}
\input{body/method/attention}
\input{body/method/treelstm}
\input{body/method/evaluation}

%% file: body/method/sentence-encoders.tex
\subsection{Priors from Random Sentence Encoders}
\label{subsec:sentence-encoders}

The line of investigation we take follows \citet{Wieting2019} closely.
We treat randomly initialized \glspl{nn} as handcrafted priors
for how the meaning of a sentence is composed from its constituent words.
Concretely, let each word $\txtw$ have a \emph{pre-trained} and
\emph{fixed} $D$-dimensional word
representation $\mbe_{\txtw}\in\mbbR^D$.
Consider a sentence $\txtS$ consisting of $\txtT_{\txtS}$ words
$\txtS=\txtw_1,\ldots,\txtw_{\txtT_{\txtS}}$.
Using an encoding function $f_{\text{enc}}$, the meaning of the sentence is
distilled into a sentence representation
$\mbh_{\txtS}$:
\begin{equation}
  \mbh_{\txtS}(\mbtheta)=f_{\text{enc}}(\bme_1,\ldots,\bme_{\txtT_{\txtS}};\mbtheta),
\end{equation}
where $\mbtheta$ are the parameters of the encoding function.
For \gls{nn} architectures that output a matrix
$\mbH_{\txtS}
\in\mbbR^{
\txtT_{\txtS}\times D^\prime}$, where $D^\prime$  is an output dimensionality and $\txtT_{\txtS}$ is a temporal
dimensionality\footnote{
  In practice $\txtT_{\txtS}$ may not directly correspond
  to the length of the input sentence due to e.g. finite kernel sizes in
  convolution operations.}, 
we pool along the temporal dimension using a pooling function $f_{\text{pool}}$.
For
our main results we use max pooling
$\mbh_{\txtS} =
f_\text{pool}(\mbH_{\txtS})=\max(\mbH_{\txtS})\in\mbbR^{D^\prime}$ throughout,
as it has been successful in InferSent \cite{Conneau2017b}.

The $\mbtheta$ are typically learned using e.g. \gls{mle} on sentence context,
resulting in $\mbh_{\txtS}(\mbtheta)$ representing a sample from the encoder's
posterior over functions applied to $\txtS$ given a corpus.
Instead of learning $\mbtheta$, we simply sample $\mbtheta$ from its own
prior. $\mbh_{\txtS}(\mbtheta)$ then represents a sample from the encoder's
prior over functions applied to $\txtS$.

For each encoding function, we take multiple samples of $\mbtheta$.
For each sample, the resulting encoder function is used to produce sentence
embeddings for a set of downstream tasks.
These downstream tasks are the supervised transfer tasks of the SentEval \cite{Conneau}
framework, where the transfer model is a simple logistic regression
model or a MLP\footnote{For emphasis: the parameters of these logistic regression model and MLP
  \emph{are} updated by the task.}.
Combining the results from multiple samples then
gives a performance estimate of each encoder's prior.

%% file: body/method/existing.tex
\subsection{BOREPs, Random LSTMs and ESNs}
\label{subsec:existing}
We take the architectures investigated in \cite{Wieting2019} as a starting
point: \gls{borep}, Random \gls{lstm} Neworks and
\glspl{esn}.
\gls{borep} is simply a random projection of word embeddings to a higher
dimension, Rand\gls{lstm} is a randomly initialised bi-directional \gls{lstm} \cite{Hochreiter1997},
and \gls{esn} is a hypertuned randomly initialised bi-directional \gls{esn} \cite{Jaeger2010}. For
more details please see \cite{Wieting2019}.

%% file: body/method/convolutional.tex
\subsection{Random \glspl{cnn}}
\label{subsec:convolutional}

\input{body/results/senteval-table}
Although \glspl{cnn} are more famously used in the image domain
\cite{Simonyan2014,He2015}, they have also enjoyed much success as sentence
encoders \cite{Collobert2011,Vieira2017,Gan2017}.
A temporal one-dimensional convolution is performed by applying a $D^\prime$-channel
filter $\bmsfW\in\mbbR^{D \times k\times D^\prime}$ to a window of $k$ words and a bias
added. This weight $\bmsfW$ is initialised uniformly at random from $ [-\frac{1}{\sqrt{d}}, \frac{1}{\sqrt{d}} ] $, where d is the word embedding dimension. The representation $\mbh_S$ is then obtained by pooling
\begin{equation}
  \mbh_{\txtS}  =f_{\text{pool}}\left[ \text{CNN}(\mbe_1,\ldots,\bme_{\txtT_{\txtS}}) \right]\in\mbbR^{D^\prime}.
\end{equation}
Note that using a window size $k=1$ corresponds to \gls{borep}.

%% file: body/results/senteval-table.tex
\renewcommand{\tabcolsep}{3pt} 
\begin{table*}[ht]
  \centering
  \small{}
  \begin{tabular}{l|l|llllllll}
    \toprule
    Model & Dim & MR & CR & MPQA & SUBJ & SST2 & TREC &   SICK-E & MRPC \\
    \midrule
    BOE$^\dagger$ & 300 & 77.3(.2) & 78.6(.3) & 87.6(.1) & 91.3(.1) & 80.0(.5) & 81.5(.8) &   78.7(.1) & 72.9(.3) \\
    \midrule
    BOREP$^\dagger$ & 4096 & 77.4(.4) & 79.5(.2) & 88.3(.2) & 91.9(.2) & 81.8(.4) & 88.8(.3) &   82.7(.7) & 73.9(.4) \\
 BOREP (ours)              & 4096 & 75.3(.2)     & 78.2(.5)     & 88.5(.2)     & 90.3(.4)     & 79.3(1.1)    & 88.5(1.3)    & 82.1(.2) & 71.8(.7) \\
    RandLSTM$^\dagger$ & 4096 & 77.2(.3) & 78.7(.5) & 87.9(.1) & 91.9(.2) & 81.5(.3) & 86.5(1.1) &   81.8(.5) & \bf 74.1(.5)\\
    RandLSTM (ours)  & 4096 &  76.9(.2) & \bf 80.9(.3) & \bf 88.7(.1) & 91.7(.1)     & 81.3(.5) & 89.2(.4)     & 81.7(.5)     & 71.8(.6)     \\
    ESN$^\dagger$ & 4096 & \bf 78.1(.3) &  80.0(.6) & 88.5(.2) & \bf 92.6(.1) & \bf 83.0(.5) & 87.9(1.0) &  \bf 83.1(.4) & 73.4(.4)\\
    ESN (ours)                  & 4096 & 70.4(.1)     & 76.9(.8)     & 86.3(.1)     & 88.7(.4)     & 76.4(.5)     & 88.9(1.2)    & 78.4(.3)     & 67.4(.7)     \\
 CNN Window = 3            & 4096 & 74.9(.3)     & 76.9(.7)     & 85.4(.2)     & 88.6(.1)     & 75.6(.5)     & 88.7(1.2)    & 79.1(.2)     & 69.4(.5)     \\
 CNN Window = 4            & 4096 & 74.3(.3)     & 74.8(.8)     & 84.2(.3)     & 86.8(.3)     & 75.5(.5)     & 85.2(1.1)    & 78.0(.2)     & 69.2(.3)     \\
 Self-Attention            & 4096 & 68.0(.3)     & 77.1(.5)     & 82.0(.5)     & 90.1(.3)     & 78.8(1.2)    & 84.9(1.3)    & 73.7(.7)     & 67.1(1.1)    \\
 TreeLSTM                  & 4096 & 75.6(.2)     & 78.5(.3)     & 87.7(.1)     & 91.4(.0)     & 79.9(.5)     & \bf 90.3(.7) & 80.7(.9)     & 71.1(.5)     \\
    \bottomrule
  \end{tabular}
  \caption{
    \label{tab:lowdimexpts}
    \small
    Performance (accuracy) for $f_{\text{pool}}=\max{}$ on all eight tasks. The results indicated by
    ${}^\dagger$ are taken from \cite{Wieting2019}. Mean (standard deviation)
    for each model is reported across five seeds. Our \gls{esn} was evaluated
    using a spectral radius of 1.0, a maximum kernel deviation from 0.0 of 0.1, and a
    sparsity 0.5, whereas the result from \cite{Wieting2019} is the best
    performing model from a hyperparameter search.}
  \label{table:results_max}
\end{table*}

%% file: body/results/max-pool-diag.tex
\begin{figure*}[ht]
  \centering
    \includegraphics[width=\linewidth]{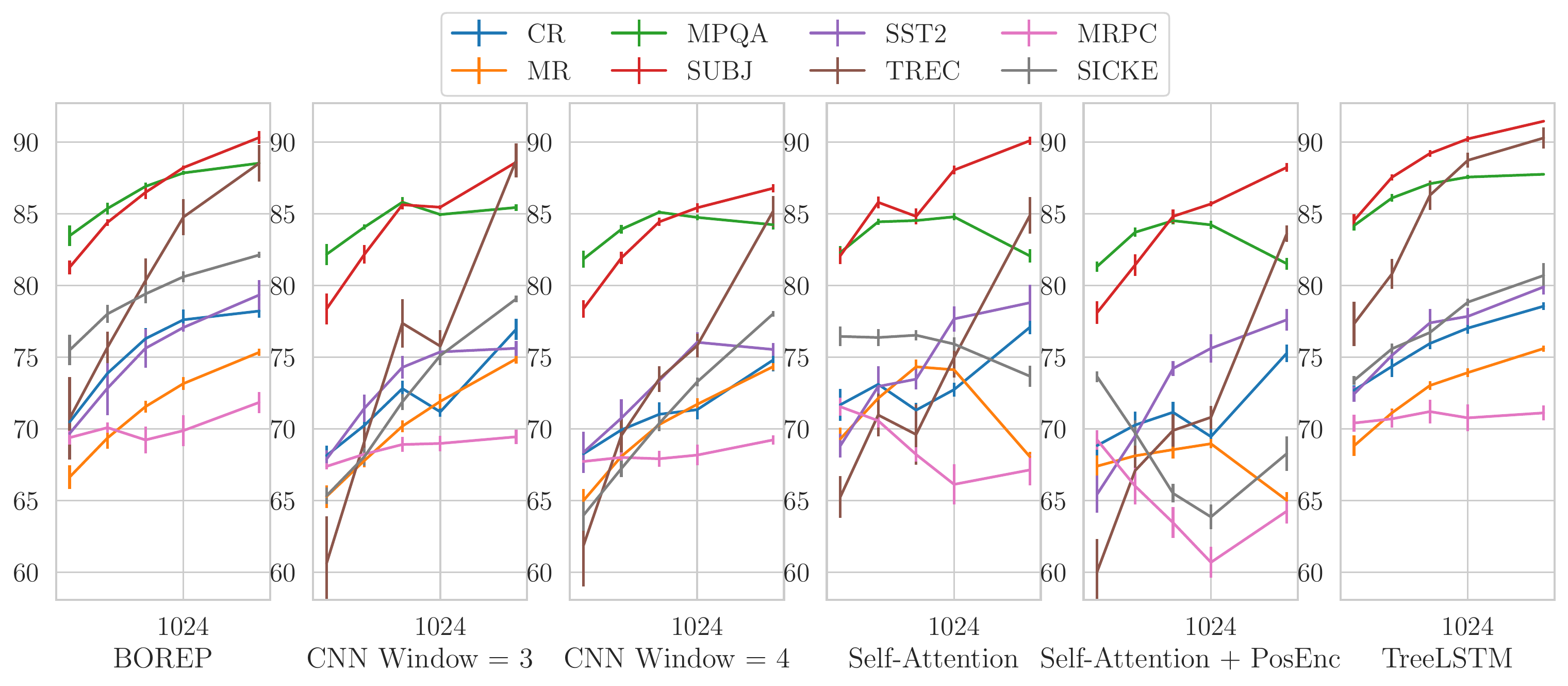}
    \caption{
      \small
      Performance (accuracy) for $f_{\text{pool}}=\max{}$ on all eight tasks
      across five seeds. We
      observe: \textbf{1)} Almost every encoder 
      architecture performs at best,
      similarly to the relatively uninformative \gls{borep}, and at worst, much
      worse. \textbf{2)}~Taking \gls{borep} as \gls{cnn}
      with a window size of 1, we note that increasing \gls{cnn} window size impairs
      performance.
      This indicates that any gains to be made from employing $n$-grams over word
      representations as a basis for distilling meaning needs to be learned. \textbf{3)}
      The performance of the
      Self-Attention Network with
      and without positional encoding is fairly similar. This indicates that
      although the encoder architecture has positional information available,
      the transfer model cannot learn to use it. It would be interesting to look at the BShift
      task to probe this directly \cite{Conneau2018a}.
      \textbf{4)} Random Self-Attention networks perform poorly even though
      they form a cornerstone of modern state of the art \gls{nlp} systems.
      Considering \Cref{eq:self-attention}, we see that the random
      contextualisation can be \emph{any} linear combination of the input, with
      none selected by an inductive bias. There is no reason to expect this
      random combination to outperform \gls{borep}.
      \textbf{5)} The Tree\gls{lstm} performs noticeably better than other encoder
  architectures on TREC, a question-type task which relies heavily on
  sentence syntax to solve \cite{Li2002a}.
  It appears that in this instance, the encoder may be using the syntactic
  information available, however, its performance on all other tasks is
  comparable to \gls{borep}.
  \label{figure:results_max}
  }
\end{figure*}

%% file: body/method/attention.tex
\subsection{Random Self-Attention}
\label{subsec:attention}
Attention mechanisms have been employed on many \gls{nlp} tasks with tremendous
success \cite{Vaswani2017, Ahmed2017a, Shaw2018, Dehghani2018a, Radford2019,
  Devlin2018, Cer2018a}. Self-attention in particular has enabled the
incorporation of incredibly long ranged contexts, as well as hierarchical
contextualisations of word embeddings within a highly parallel setting.

In our random setting, the word embeddings $\mbe_1,\ldots,\mbe_{\txtT_{\txtS}}$ are first projected up to a
$D^\prime$ dimensional space. We then optionally add sinusoidal positional
encodings \cite{Vaswani2017}.
We then apply two layers of random self-attention with residual connections, each
followed by layer normalisation.
A single head of a self-attention layer produces new embeddings for each query
representations $\mbq\in\mbbR^{d_k}$ out of the value representations
$\mbv_i\in\mbbR^{D^\prime}$, controlled by the key representations
$\mbk_i\in\mbbR^{d_k}$
\begin{equation}
  \label{eq:self-attention}
  \mbq^\prime = \sum_{i=1}^{\txtT_{\txtS}} \exp\left(\mbq^T\,\mbk_i/\sqrt{d_k}\right)\,\mbv_i / \text{constant}.
\end{equation} 
The $d_k$-dimensional key and query representations are given by independent random
projections acting upon the self-attention layer input. We use eight heads of
attention in each layer. The pooling function is applied to this output to
produce the sentence representation $\mbh_{\txtS}$.

We keep the default initialisation of the FairSeq implementation, which is
Xavier uniform \cite{Glorot2010} for the weights of the self-attention layer.

%% file: body/method/treelstm.tex
\subsection{Random Tree\glspl{lstm}}
\label{subsec:treelstm}
The final architecture we consider is the Tree\gls{lstm}. This
architecture is particularly interesting as it can potentially incorporate
syntactic information into the sentence representations \cite{Tai2015, Li,
  Zhang, Teng2016, Kim2018, Ahmed2019}.

We specifically consider the
Binary Constituency Tree\gls{lstm} \cite{Tai2015}. This differs from a regular
\gls{lstm} by having a two forget gates - one for each child node given by the
structure of the parsed sentence.

Word representations are first presented to a random bi-directional \gls{lstm} of
combined dimensionality $D^\prime$ to provide contextualised representations
$\mbE^\prime_\txtS\in\mbbR^{\txtT_{\txtS}\times D}$
\begin{equation}
  \mbE^\prime_{\txtS} = \text{BiLSTM}(\mbe_1,\ldots,\mbe_{\txtT_{\txtS}}).
\end{equation}
The contextualised representations are then presented to a random Tree\gls{lstm},
whose outputs are pooled to produce the sentence representation
\begin{equation}
\mbh_{\txtS} = f_{\text{pool}}\left[\text{TreeLSTM}(\mbE^\prime_{\txtS})\right].
\end{equation}
Both weights of the bi-directional \gls{lstm} and the Tree\gls{lstm} are initialised uniformly at random from $ [-\frac{1}{\sqrt{d}}, \frac{1}{\sqrt{d}} ] $. We used the Stanford parser \cite{Manning2015} to parse each sentence.
Punctuation and special characters were removed, and numbers were only kept if they
formed an independent word and were not part of a mixed word of letters and
numbers. Then, in the length of a
word was reduced to zero, the word was replaced with a placeholder \texttt{*}
character.
After parsing,
the prepossessing described in \cite{Kim2018} was used to compute the parse
tree for the Tree\gls{lstm}.

%% file: body/method/evaluation.tex
\subsection{Evaluation}
\label{subsec:evaluation}
The SentEval tasks we evaluate on are sentiment analysis (MR,
SST), question-type (TREC), product review (CR), subjectivity (SUBJ), opinion
polarity (MPQA), paraphrasing (MRPC), and entailment (SICK-E). We use the
default SentEval settings defined in \cite{Conneau}.
We evaluate for five samples (seeds) per architecture per task.

We follow the FairSeq implementation \cite{Ott2019} to build our \gls{cnn} and
self-attention networks. We also follow the implementation of \cite{Kim2018a}
without the structure-aware tag representations to build our Tree\glspl{lstm}.

%% file: body/results.tex
\section{Results}
\label{sec:results}
Our investigation is concerned with the priors of encoder architectures, rather
than the posteriors they may learn from data;
we only compare
untrained encoders acting upon word embeddings.

\Cref{table:results_max} contains the performance of architectures discussed in
\Cref{sec:method} at dimensionality 4096 on the selected SentEval tasks,
together with the results from \citet{Wieting2019}. \Cref{figure:results_max}
contains the performance for these architectures across a range of
dimensionalities.

As a sanity check, we evaluated \gls{borep} and \gls{cnn} with a window size of
1 and found the performance indistinguishable.

In general, we find that even if encoders have inductive biases that present
additional information that can be used to solve a task, the corresponding priors
do not leverage this information, except in an isolated case.
This strongly indicates that learning is an essential component of building
encoder architectures if any gains are to be made beyond apparently
uninformative priors. 

%% file: body/conclusion.tex
\section{Conclusion}
\label{sec:conclusion}
We have evaluated randomly initialised architectures
to measure the contribution of priors in distilling sentence meaning.
We find that apparently uninformative priors are just as good as seemingly
informative priors on almost all tasks, indicating that learning is a necessary
component to leverage information provided by architecture choice.

%% file: body/acknowledgements.tex
\section{Acknowledgements}
\label{sec:acknowledgements}
We thank Jeremie Vallee for assistance with the experimental setup, and the wider machine learning group at Babylon for useful comments and support throughout this project.

%% file: bibliography.tex
\bibliographystyle{acl_natbib}

\input{ms.bbl}

%% file: ms.bbl
\begin{thebibliography}{43}
\expandafter\ifx\csname natexlab\endcsname\relax\def\natexlab#1{#1}\fi

\bibitem[{Ahmed et~al.(2017)Ahmed, Keskar, and Socher}]{Ahmed2017a}
Karim Ahmed, Nitish~Shirish Keskar, and Richard Socher. 2017.
\newblock \href {http://arxiv.org/abs/1711.02132} {{Weighted Transformer
  Network for Machine Translation}}.
\newblock pages 1--10.

\bibitem[{Ahmed et~al.(2019)Ahmed, Samee, and Mercer}]{Ahmed2019}
Mahtab Ahmed, Muhammad~Rifayat Samee, and Robert~E. Mercer. 2019.
\newblock \href {https://doi.org/10.1109/ICOSC.2019.8665673} {{Improving
  Tree-LSTM with Tree Attention}}.
\newblock In \emph{2019 IEEE 13th International Conference on Semantic
  Computing (ICSC)}, pages 247--254. IEEE.

\bibitem[{Ba et~al.(2016)Ba, Kiros, and Hinton}]{Ba2016}
Jimmy~Lei Ba, Ryan Kiros, and Geoffrey~E. Hinton. 2016.
\newblock \href {http://arxiv.org/abs/1607.06450} {{Layer Normalization}}.

\bibitem[{Bastings et~al.(2017)Bastings, Titov, Aziz, Marcheggiani, and
  Sima'an}]{Bastings2017}
Joost Bastings, Ivan Titov, Wilker Aziz, Diego Marcheggiani, and Khalil
  Sima'an. 2017.
\newblock \href {http://arxiv.org/abs/1704.04675} {{Graph Convolutional
  Encoders for Syntax-aware Neural Machine Translation}}.
\newblock pages 1957--1967.

\bibitem[{Battaglia et~al.(2018)Battaglia, Hamrick, Bapst, Sanchez-Gonzalez,
  Zambaldi, Malinowski, Tacchetti, Raposo, Santoro, Faulkner, Gulcehre, Song,
  Ballard, Gilmer, Dahl, Vaswani, Allen, Nash, Langston, Dyer, Heess, Wierstra,
  Kohli, Botvinick, Vinyals, Li, and Pascanu}]{Authors2018}
Peter~W. Battaglia, Jessica~B. Hamrick, Victor Bapst, Alvaro Sanchez-Gonzalez,
  Vinicius Zambaldi, Mateusz Malinowski, Andrea Tacchetti, David Raposo, Adam
  Santoro, Ryan Faulkner, Caglar Gulcehre, Francis Song, Andrew Ballard, Justin
  Gilmer, George Dahl, Ashish Vaswani, Kelsey Allen, Charles Nash, Victoria
  Langston, Chris Dyer, Nicolas Heess, Daan Wierstra, Pushmeet Kohli, Matt
  Botvinick, Oriol Vinyals, Yujia Li, and Razvan Pascanu. 2018.
\newblock \href {http://arxiv.org/abs/1806.01261} {{Relational inductive
  biases, deep learning, and graph networks}}.
\newblock \emph{Under Review}, pages 1--37.

\bibitem[{Cer et~al.(2018)Cer, Yang, Kong, Hua, Limtiaco, John, Constant,
  Guajardo-Cespedes, Yuan, Tar, Sung, Strope, and Kurzweil}]{Cer2018a}
Daniel Cer, Yinfei Yang, Sheng-yi Kong, Nan Hua, Nicole Limtiaco, Rhomni~St.
  John, Noah Constant, Mario Guajardo-Cespedes, Steve Yuan, Chris Tar,
  Yun-Hsuan Sung, Brian Strope, and Ray Kurzweil. 2018.
\newblock \href {http://arxiv.org/abs/1803.11175} {{Universal Sentence
  Encoder}}.

\bibitem[{Collobert et~al.(2011)Collobert, Weston, Bottou, Karlen, Kavukcuoglu,
  and Kuksa}]{Collobert2011}
Ronan Collobert, Jason Weston, Leon Bottou, Michael Karlen, Koray Kavukcuoglu,
  and Pavel Kuksa. 2011.
\newblock \href {http://arxiv.org/abs/1103.0398} {{Natural Language Processing
  (almost) from Scratch}}.

\bibitem[{Conneau and Kiela(2018)}]{Conneau}
Alexis Conneau and Douwe Kiela. 2018.
\newblock \href {http://arxiv.org/abs/1803.05449} {{SentEval: An Evaluation
  Toolkit for Universal Sentence Representations}}.

\bibitem[{Conneau et~al.(2017)Conneau, Kiela, Schwenk, Barrault, and
  Bordes}]{Conneau2017b}
Alexis Conneau, Douwe Kiela, Holger Schwenk, Loic Barrault, and Antoine Bordes.
  2017.
\newblock \href {http://arxiv.org/abs/1705.02364} {{Supervised Learning of
  Universal Sentence Representations from Natural Language Inference Data}}.

\bibitem[{Conneau et~al.(2018)Conneau, Kruszewski, Lample, Barrault, and
  Baroni}]{Conneau2018a}
Alexis Conneau, German Kruszewski, Guillaume Lample, Lo{\"{i}}c Barrault, and
  Marco Baroni. 2018.
\newblock \href {http://arxiv.org/abs/1805.01070} {{What you can cram into a
  single vector: Probing sentence embeddings for linguistic properties}}.

\bibitem[{Dehghani et~al.(2018)Dehghani, Gouws, Vinyals, Uszkoreit, and
  Kaiser}]{Dehghani2018a}
Mostafa Dehghani, Stephan Gouws, Oriol Vinyals, Jakob Uszkoreit, and {\L}ukasz
  Kaiser. 2018.
\newblock \href {http://arxiv.org/abs/1807.03819} {{Universal Transformers}}.
\newblock pages 1--23.

\bibitem[{Devlin et~al.(2018)Devlin, Chang, Lee, and Toutanova}]{Devlin2018}
Jacob Devlin, Ming-Wei Chang, Kenton Lee, and Kristina Toutanova. 2018.
\newblock \href {http://arxiv.org/abs/1810.04805} {{BERT: Pre-training of Deep
  Bidirectional Transformers for Language Understanding}}.

\bibitem[{Gan et~al.(2016)Gan, Pu, Henao, Li, He, and Carin}]{Gan2017}
Zhe Gan, Yunchen Pu, Ricardo Henao, Chunyuan Li, Xiaodong He, and Lawrence
  Carin. 2016.
\newblock \href {http://arxiv.org/abs/1611.07897} {{Learning Generic Sentence
  Representations Using Convolutional Neural Networks}}.
\newblock \emph{Emnlp}, pages 2380--2390.

\bibitem[{Geman et~al.(2008)Geman, Bienenstock, and Doursat}]{Geman2008}
Stuart Geman, Elie Bienenstock, and Ren{\'{e}} Doursat. 2008.
\newblock \href {https://doi.org/10.1162/neco.1992.4.1.1} {{Neural Networks and
  the Bias/Variance Dilemma}}.
\newblock \emph{Neural Computation}, 4(1):1--58.

\bibitem[{Glorot and Bengio(2010)}]{Glorot2010}
Xavier Glorot and Yoshua Bengio. 2010.
\newblock \href {https://doi.org/10.1.1.207.2059} {{Understanding the
  difficulty of training deep feedforward neural networks}}.
\newblock \emph{Proceedings of the Thirteenth International Conference on
  Artificial Intelligence and Statistics (AISTATS-10)}, 9:249--256.

\bibitem[{Harris(1954)}]{Harris1954}
Zellig~S. Harris. 1954.
\newblock \href {https://doi.org/10.1080/00437956.1954.11659520}
  {{Distributional Structure}}.
\newblock \emph{WORD}, 10(2-3):146--162.

\bibitem[{He et~al.(2015)He, Zhang, Ren, and Sun}]{He2015}
Kaiming He, Xiangyu Zhang, Shaoqing Ren, and Jian Sun. 2015.
\newblock \href {http://arxiv.org/abs/1512.03385} {{Deep Residual Learning for
  Image Recognition}}.

\bibitem[{Hochreiter and Schmidhuber(1997)}]{Hochreiter1997}
Sepp Hochreiter and J{\"{u}}rgen Schmidhuber. 1997.
\newblock \href {https://doi.org/10.1162/neco.1997.9.8.1735} {{Long Short-Term
  Memory}}.
\newblock \emph{Neural Computation}, 9(8):1735--1780.

\bibitem[{Jaeger(2001)}]{Jaeger2010}
Herbert Jaeger. 2001.
\newblock {The “echo state” approach to analysing and training recurrent
  neural networks - with an Erratum note}.
\newblock Technical Report 148.

\bibitem[{Kim et~al.(2018{\natexlab{a}})Kim, Lee, Song, and Yoon}]{Kim2018}
Sungwon Kim, Sang-gil Lee, Jongyoon Song, and Sungroh Yoon. 2018{\natexlab{a}}.
\newblock \href {http://arxiv.org/abs/1811.02155} {{FloWaveNet : A Generative
  Flow for Raw Audio}}.

\bibitem[{Kim et~al.(2018{\natexlab{b}})Kim, Choi, Edmiston, Bae, and
  Lee}]{Kim2018a}
Taeuk Kim, Jihun Choi, Daniel Edmiston, Sanghwan Bae, and Sang-goo Lee.
  2018{\natexlab{b}}.
\newblock \href {http://arxiv.org/abs/1809.02286} {{Dynamic Compositionality in
  Recursive Neural Networks with Structure-aware Tag Representations}}.

\bibitem[{Kiros et~al.(2015)Kiros, Zhu, Salakhutdinov, Zemel, Torralba,
  Urtasun, and Fidler}]{Kiros2015}
Ryan Kiros, Yukun Zhu, Ruslan Salakhutdinov, Richard~S. Zemel, Antonio
  Torralba, Raquel Urtasun, and Sanja Fidler. 2015.
\newblock \href {http://arxiv.org/abs/1506.06726} {{Skip-Thought Vectors}}.

\bibitem[{Lempitsky et~al.(2018)Lempitsky, Vedaldi, and
  Ulyanov}]{Lempitsky2018}
Victor Lempitsky, Andrea Vedaldi, and Dmitry Ulyanov. 2018.
\newblock \href {https://doi.org/10.1109/CVPR.2018.00984} {{Deep Image Prior}}.
\newblock In \emph{2018 IEEE/CVF Conference on Computer Vision and Pattern
  Recognition}, pages 9446--9454. IEEE.

\bibitem[{Li and Roth(2002)}]{Li2002a}
Xin Li and Dan Roth. 2002.
\newblock \href {https://doi.org/10.3115/1072228.1072378} {{Learning question
  classifiers}}.
\newblock In \emph{Proceedings of the 19th international conference on
  Computational linguistics -}, volume~1, pages 1--7, Morristown, NJ, USA.
  Association for Computational Linguistics.

\bibitem[{Li et~al.(2018)Li, Richtarik, Ding, and Gao}]{Li}
Yu~Li, Peter Richtarik, Lizhong Ding, and Xin Gao. 2018.
\newblock \href {http://arxiv.org/abs/1808.05385} {{On the Decision Boundary of
  Deep Neural Networks}}.

\bibitem[{Manning et~al.(2014)Manning, Surdeanu, Bauer, Finkel, Bethard, and
  McClosky}]{Manning2015}
Christopher Manning, Mihai Surdeanu, John Bauer, Jenny Finkel, Steven Bethard,
  and David McClosky. 2014.
\newblock \href {https://doi.org/10.3115/v1/P14-5010} {{The Stanford CoreNLP
  Natural Language Processing Toolkit}}.
\newblock In \emph{Proceedings of 52nd Annual Meeting of the Association for
  Computational Linguistics: System Demonstrations}, pages 55--60, Stroudsburg,
  PA, USA. Association for Computational Linguistics.

\bibitem[{Manning et~al.(2008)Manning, Raghavan, and Schutze}]{Manning2008}
Christopher~D. Manning, Prabhakar Raghavan, and Hinrich Schutze. 2008.
\newblock \href {https://doi.org/10.1017/CBO9780511809071} {\emph{{Introduction
  to Information Retrieval}}}.

\bibitem[{Marcheggiani et~al.(2018)Marcheggiani, Bastings, and
  Titov}]{Marcheggiani2018}
Diego Marcheggiani, Joost Bastings, and Ivan Titov. 2018.
\newblock \href {http://arxiv.org/abs/1804.08313} {{Exploiting Semantics in
  Neural Machine Translation with Graph Convolutional Networks}}.

\bibitem[{Marcheggiani and Perez-Beltrachini(2018)}]{Marcheggiani2018a}
Diego Marcheggiani and Laura Perez-Beltrachini. 2018.
\newblock \href {http://arxiv.org/abs/1810.09995} {{Deep Graph Convolutional
  Encoders for Structured Data to Text Generation}}.
\newblock pages 1--9.

\bibitem[{Marcheggiani and Titov(2017)}]{Marcheggiani2017a}
Diego Marcheggiani and Ivan Titov. 2017.
\newblock \href {http://arxiv.org/abs/1703.04826} {{Encoding Sentences with
  Graph Convolutional Networks for Semantic Role Labeling}}.
\newblock pages 1506--1515.

\bibitem[{Mitchell(1991)}]{Mitchell1980}
Tom Mitchell. 1991.
\newblock {The need for biases in learning generalisations}.
\newblock \emph{Readings in Machine Learning}, (May).

\bibitem[{Ott et~al.(2019)Ott, Edunov, Baevski, Fan, Gross, Ng, Grangier, and
  Auli}]{Ott2019}
Myle Ott, Sergey Edunov, Alexei Baevski, Angela Fan, Sam Gross, Nathan Ng,
  David Grangier, and Michael Auli. 2019.
\newblock \href {http://arxiv.org/abs/1904.01038} {{fairseq: A Fast, Extensible
  Toolkit for Sequence Modeling}}.

\bibitem[{Radford et~al.(2019)Radford, Wu, Child, Luan, Amodei, and
  Sutskever}]{Radford2019}
Alec Radford, Jeffrey Wu, Rewon Child, David Luan, Dario Amodei, and Ilya
  Sutskever. 2019.
\newblock \href {https://github.com/codelucas/newspaper} {{Language Models are
  Unsupervised Multitask Learners}}.
\newblock \emph{Open AI}.

\bibitem[{Salton et~al.(1975)Salton, Wong, and Yang}]{Salton1975}
G.~Salton, A.~Wong, and C.~S. Yang. 1975.
\newblock \href {https://doi.org/10.1145/361219.361220} {{A vector space model
  for automatic indexing}}.
\newblock \emph{Communications of the ACM}, 18(11):613--620.

\bibitem[{Shaw et~al.(2018)Shaw, Uszkoreit, and Vaswani}]{Shaw2018}
Peter Shaw, Jakob Uszkoreit, and Ashish Vaswani. 2018.
\newblock \href {http://arxiv.org/abs/1803.02155} {{Self-Attention with
  Relative Position Representations}}.

\bibitem[{Simonyan and Zisserman(2014)}]{Simonyan2014}
Karen Simonyan and Andrew Zisserman. 2014.
\newblock \href {http://arxiv.org/abs/1409.1556} {{Very Deep Convolutional
  Networks for Large-Scale Image Recognition}}.
\newblock pages 1--14.

\bibitem[{Tai et~al.(2015)Tai, Socher, and Manning}]{Tai2015}
Kai~Sheng Tai, Richard Socher, and Christopher~D. Manning. 2015.
\newblock \href {http://arxiv.org/abs/1503.00075} {{Improved Semantic
  Representations From Tree-Structured Long Short-Term Memory Networks}}.

\bibitem[{Teng and Zhang(2016)}]{Teng2016}
Zhiyang Teng and Yue Zhang. 2016.
\newblock \href {http://arxiv.org/abs/1611.06788} {{Bidirectional
  Tree-Structured LSTM with Head Lexicalization}}.

\bibitem[{Vaswani et~al.(2017)Vaswani, Shazeer, Parmar, Uszkoreit, Jones,
  Gomez, Kaiser, and Polosukhin}]{Vaswani2017}
Ashish Vaswani, Noam Shazeer, Niki Parmar, Jakob Uszkoreit, Llion Jones,
  Aidan~N. Gomez, Lukasz Kaiser, and Illia Polosukhin. 2017.
\newblock \href {http://arxiv.org/abs/1706.03762} {{Attention Is All You
  Need}}.

\bibitem[{Vieira and Moura(2017)}]{Vieira2017}
Joao Paulo~Albuquerque Vieira and Raimundo~Santos Moura. 2017.
\newblock \href {https://doi.org/10.1109/CLEI.2017.8226381} {{An analysis of
  convolutional neural networks for sentence classification}}.
\newblock In \emph{2017 XLIII Latin American Computer Conference (CLEI)},
  volume 2017-Janua, pages 1--5. IEEE.

\bibitem[{Wieting and Kiela(2019)}]{Wieting2019}
John Wieting and Douwe Kiela. 2019.
\newblock \href {http://arxiv.org/abs/1901.10444} {{No Training Required:
  Exploring Random Encoders for Sentence Classification}}.
\newblock pages 1--16.

\bibitem[{Zhang et~al.(2016)Zhang, Bengio, Hardt, Recht, and
  Vinyals}]{Zhang2016a}
Chiyuan Zhang, Samy Bengio, Moritz Hardt, Benjamin Recht, and Oriol Vinyals.
  2016.
\newblock \href {https://doi.org/10.1109/TKDE.2015.2507132} {{Understanding
  deep learning requires rethinking generalization}}.

\bibitem[{Zhang et~al.(2019)Zhang, Bengio, Hardt, and Singer}]{Zhang}
Chiyuan Zhang, Samy Bengio, Moritz Hardt, and Yoram Singer. 2019.
\newblock \href {http://arxiv.org/abs/1902.04698} {{Identity Crisis:
  Memorization and Generalization under Extreme Overparameterization}}.
\newblock pages 1--28.

\end{thebibliography}
